\documentclass[conference]{IEEEtran}
\usepackage{stmaryrd}
\usepackage{amsfonts}
\usepackage{graphicx}
\usepackage{times}
\usepackage{cite}
\usepackage[cmex10]{amsmath}
\usepackage{ulsy}
\usepackage{url}
\usepackage{color}

\usepackage[utf8]{inputenc}
\usepackage{authblk}

\IEEEoverridecommandlockouts    

\begin{document}

\title{\ \\ \LARGE\bf Autonomous Reinforcement of Behavioral Sequences \\ in Neural Dynamics \thanks{S.K. and M.L. are with the Istituto Dalle Molle di Studi sull'Intelligenza Artificiale (IDSIA), Manno-Lugano, Switzerland (email: \{sohrob, matthew\}@idsia.ch).  M.R. and Y.S. are with the Institut f\"ur Neuroinformatik at the Universit\"atstr, Bochum, Germany (email: \{mathis.richter, yulia.sandamirskaya\}@ini.ruhr-uni-bochum.de)} \thanks{This work was supported in the European Union Seventh Framework Program FP7-ICT-2009-6 under Grant Agreement no. 270247 -- NeuralDynamics.}}

\author[$\dagger$]{Sohrob Kazerounian}
\author[$\dagger$]{Matthew Luciw}
\author[ ]{Mathis Richter}
\author[ ]{Yulia Sandamirskaya}
\affil[$\dagger$]{Joint first authors.}
\renewcommand\Authands{ and }


\maketitle

\begin{abstract}
We introduce a dynamic neural algorithm called Dynamic Neural (DN) SARSA($\lambda$) for learning a behavioral sequence from delayed reward. DN-SARSA($\lambda$) combines Dynamic Field Theory models of behavioral sequence representation, classical reinforcement learning, and a computational neuroscience model of working memory, called Item and Order working memory, which serves as an eligibility trace.  DN-SARSA($\lambda$) is implemented on both a simulated and real robot that must learn a specific rewarding sequence of elementary behaviors from exploration.  Results show DN-SARSA($\lambda$) performs on the level of the discrete SARSA($\lambda$), validating the feasibility of general reinforcement learning without compromising neural dynamics.
\end{abstract}

\section{Introduction}

\textbf{Standard approaches to reinforcement learning} (RL;~\cite{sutton1998reinforcement}) typically formalize the learning problem in terms of discrete state and action spaces, and have a learning agent that operates in discrete time.  How can such discrete representations emerge from spatiotemporally continuous sensory and motor representations?  \textbf{Computational neuroscience models of learning from reward} do include continuous neural representations of states and actions, but these typically involve purely \textit{immediate} rewards; thus not addressing the general RL problem, in which the reward is potentially delayed.  How can a neural model learn rewarding sequences through delayed reinforcement, and, further, how can these sequences be generated using real sensors and motors~\cite{schultz1997neural,berns1998computational}?

To understand the contribution herein, one needs to understand three perspectives: at the core is the \textbf{Dynamic Field Theory} (DFT) perspective.  DFT has not yet integrated general learning from reward into its framework. Our purpose is to add RL to DFT, but \textbf{standard reinforcement learning} is not completely appropriate for DFT, while \textbf{computational neuroscience RL} does not address the general learning problem, nor is it typically fast enough for real-time robotic systems.  We arrive at a middle ground: an algorithm we call Dynamic-Neural (DN) SARSA($\lambda$).  This \textit{neural-dynamic} RL model is able to learn action sequences from a delayed reward signal over state and action representations that are continuously linked to raw perceptual inputs and motor dynamics.  Moreover, the model shows how eligibility traces (ET), can be realized in neural circuits by implementing the ET as an Item and Order working memory\cite{grossberg2011neural, grossberg1978WM}.

\section{Background}
\subsection{Dynamic Field Theory Perspective}
Dynamic Field Theory (DFT~\cite{Schoner2008}) is a mathematical framework, originally used to model reactive motor behaviors~\cite{kopecz1995saccadic}, which, more recently, has been used to model complex cognitive processes~\cite{spencer2009dynamic}. In DFT, dynamic neural fields (DNFs) represent activation distributions of neural populations. Activation is over graded metric dimensions (e.g., color or space) and develops in continuous time based on the classical Amari dynamics~\cite{Amari77}. Stable \emph{peaks} of activation form as a result of supra-threshold activation and lateral interactions within a field.  DFT architectures are able to deal with continuous time and real world environments and are thus well suited for robotic control systems.

The basic learning mechanism in DFT has been a memory trace of the positive activation of a DNF. This mechanism has shown quite flexible: it has been used to model long-term memory with respect to task space~\cite{KopeczSchoner95,ErlhagenSchoner2002}, the motor memory of previous movements~\cite{ThelenEtAl2001,SchonerThelen2006}, to encode invariant features~\cite{faubel2008learning}, and to represent locations of objects~\cite{ZibnerEtAl2010ICDL}. In these models, learning is achieved by the dynamics of the memory trace's build-up and decay.  Memory traces of multi-dimensional DNFs implement associative learning between different modalities. These associations have been used to learn a serial order of actions~\cite{sandamirskaya2010embodied}, from teacher supervision.  We would like  the rewarding sequence of behaviors to be learned autonomously based on a delayed and non-specific reward signal.   While it is impressive what has been accomplished using the memory trace alone,  it is insufficient for this purpose.

\subsection{Standard RL Perspective}

Standard RL algorithms~\cite{sutton1998reinforcement} bring strong guarantees of convergence and optimality for any type of reward function.  A general statement of the RL problem starts with an agent that interacts with its environment. At any moment in discrete time ($t$), the agent observes the state of the environment ($s_t$), then makes a decision about which action to take ($a_t$). The decision is determined by the agent's policy ($\pi (s,a)$).  The learning task is to attain the optimal policy, which, when followed, maximizes the long-term reward, through exploration.  Value-based RL methods focus on estimating the value function of the optimal policy (the value of a state-action is the expected future cumulative discounted reward if the agent takes that action in that state and follows its policy thereafter).

At the heart of standard value-based RL is Temporal-Difference (TD) learning. During exploration, TD learning updates the estimated value of a particular state (or state-action), based on the estimated value of the subsequent state. The SARSA algorithm makes use of TD learning to incrementally update state-action values of the agent's exploration policy.  Coupled with policy improvement (by taking the action with the highest estimated value), SARSA will converge to the optimal policy \textit{for any reward function}~\cite{sutton1998reinforcement}.  SARSA($\lambda$) introduces the eligibility trace (tuned by $0 < \lambda \le 1$), so that not just the previous state-action value is updated, but a limited history of state-action values. That is, if we denote our TD-error at any given time $t$ as $\delta_t$, state-action values which occurred $t$ timesteps back, are updated by a factor of $\gamma^t \lambda \delta_t$, where $0 < \gamma \le 1$ is the discount factor.

In many formulations, the environment is a Markov Decision Process (MDP), i.e., the response of the environment depends solely on the current state and action (the history of states or actions prior is irrelevant). An example would be a game of chess, where the next board configuration depends only on the current configuration and the selected action, rather than the sequence of moves which lead to that configuration. But in many environments this property does not hold. In these cases, the environment is a partially observable MDP (POMDP).   The use of eligibility traces has not only been shown to speed up learning, but also been shown to help overcome the problem of learning in POMDPs~\cite{todd2009learning}.

\subsection{Computational Neuroscience RL Perspective}

A number of neural models have been able to model low-level aspects of reinforcement learning, including sequence production in Basal Ganglia~\cite{berns1998computational}, foraging behavior in bees~\cite{montague1995bee}, and planned and reactive saccades~\cite{brown2004laminar}. However, while these models explain an impressive array of physiological data regarding RL, they make simplified assumptions about the nature of the environments they model, mainly dealing with purely immediate reward in a basic conditioning sense.  Moreover, they do not account for how behavior is generated in continuous time based on realistic sensory information and tied into actual motor systems.

As noted by Kawato and Sanejima ~\cite{kawato2007efficient}, there are three primary problems facing neural models of RL. First, the neural TD algorithms learn too slowly to be considered realistic methods of learning, either in animals or in robots. Second, even though there is accumulating neurophysiological evidence that midbrain dopaminergic neurons encode TD-error~\cite{schultz1997neural}, the exact mechanisms by which TD-errors are computed by neural circuits remain elusive.  Specifically, the TD error involving value estimates alone, in the case of non-immediate reward, is not accounted for in existing models.  Third, neural models of RL fail to explain complex behavioral learning which incorporate cerebral cortex and cerebellum.

DN-SARSA($\lambda$) provides a framework which can address these conceptual issues, by showing how computational enhancements to learning, such as eligibility traces, can be realized in neural circuits; to propose a mechanism by which TD-errors with eligibility traces can be computed, while maintaining the Bellman consistency; and to show how neural reinforcement learning algorithms can interact with sensory cortices, all of which operate in real-time, on real inputs.  Our overall architecture therefore integrates standard RL, with its strong guarantees for general environment, and neuroscience-based RL, with its biological plausibility for continuous environment, with DFT, with its capacity to handle complex, real-time and dynamic environments.

\section{The DN-SARSA($\lambda$) Architecture}
\label{sec:architecture}

\begin{figure}[!t]
\centering
\includegraphics[width=0.46\textwidth]{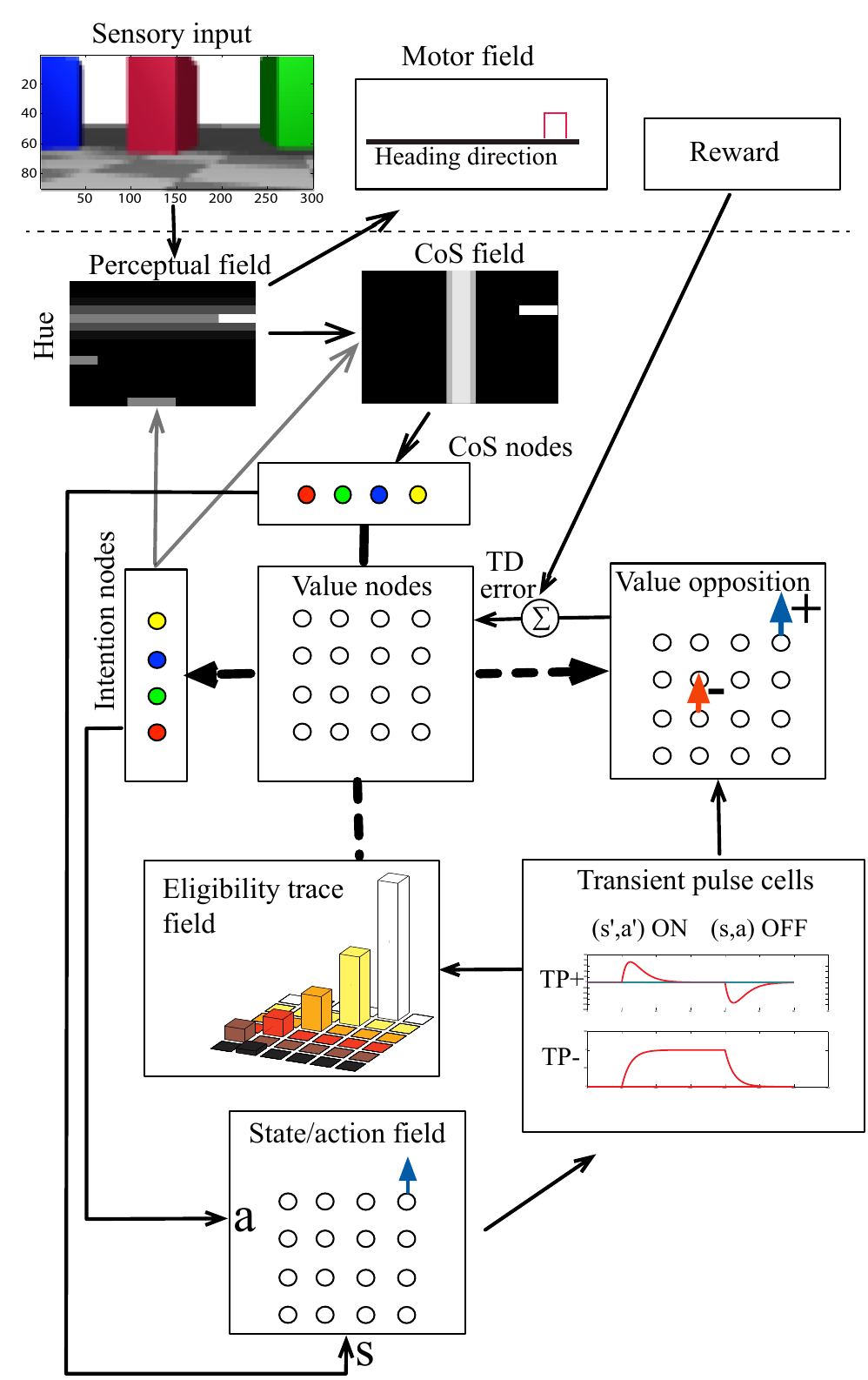}
\caption{System Architecture.  See text for details.}
\label{FG:ARCH1}
\end{figure}

\subsection{Overview}

The DN-SARSA($\lambda$) model consists of a neural-dynamic architecture for generation of behavioral sequences as well as a neural-dynamic reinforcement learner that learns the values of each behavior, relative to a behavior that preceded it.  Fig.~\ref{FG:ARCH1} shows a diagram of an architecture, which includes DN-SARSA($\lambda$) and simplified sensory and motor systems.  \textbf{Elementary behaviors.}  A number of coupled dynamic neural fields (DNFs)~\cite{Amari77}  and neural nodes form the elementary behaviors (EBs) of the agent's behavioral repertoire.  Each EB is defined by an intention and the condition-of-satisfaction (CoS).  The \textbf{intention node} has a weight vector that biases (e.g., selects) specific bottom-up sensory inputs (detected features from the environment) and uses the information gathered from the selected features to drive low-level motor commands.  For example, an intention of a ``go to red'' behavior biases the red hue in the input color space, so that only the position of the red object becomes salient.  This position becomes an attractor to a motor system, and the robot will go towards the red object.  The corresponding \textbf{CoS node} has an input bias that is used to tell when the behavior has successfully completed.  This is similar to~\cite{sandamirskaya2011neural}, where a sequence of behaviors is chained together through a set of ordinal nodes; instead we want to learn a particular sequence from delayed reward.

For the reinforcement learner, an active CoS node represents the \emph{state}, following which the agent decides an \emph{action} --- which intention to activate next. A \textbf{state-action DNF} builds a peak of positive activation in each transition phase between EBs, when the CoS of the previous EB is still active and the intention of the next EB is already activated.

The positive activation in the state-action DNF ultimately serves as input to an \textbf{Item-and-Order working memory}~\cite{grossberg1978WM, grossbergMemTheory}, wherein the order of a sequence is encoded by relative amounts of activity across neurons representing these transitions. We argue that the properties of Item and Order working memories which give rise to this relative (graded) activity pattern, are also what make it computationally analogous to an eligibility trace, and are crucial in their ability to maintain fixed (sustained) activation levels during variably timed action intervals while an action is being produced.

The eligibility trace's pattern of activity excites a \textbf{value opposition (VO) field}, which sets input to a dynamical array performing calculation of a Temporal Difference error. The calculated value of the TD-error modulates learning, implemented as a Hebb-like process,  whose long-term memory values, i.e., \textbf{value nodes}, represent the stored values (state-action, or Q, values) of the reinforcement learner. The Q-values are updated in the learning process and are utilized during sequence production to select the next EB.

\subsection{Sequence Generation Dynamics}

\textbf{Dynamic Neural Fields.}  The activation level of DNFs uses the following differential equation, as analyzed by \cite{Amari77}
\begin{equation}
  \label{eq:apdx:neural_field}
  \tau \dot u(x,t) = -u(x,t) + h + S(x,t) + \int \omega(x - x') \sigma(u(x,t)) dx',
\end{equation}
where $h < 0$ is a negative resting level and $S(x,t)$ is the sum of external inputs, for instance from sensors or other DNFs. The Gaussian-shaped kernel $\omega(\Delta x)$ determines the lateral interaction within the field. For supra-threshold activation, this interaction leads to stable peaks of activation, the unit of representation in DFT.

\textbf{Elementary Behaviors.}  In order to represent  actions (e.g., "move to red object") in a real-world environment and in continuous time, we use a DFT based model of an \emph{elementary behavior} \cite{SandamirskayaRichterSchoner2011}. An EB consists of two dynamical structures: a representation of the \emph{intention} (e.g., move toward red object) and of the \emph{condition of satisfaction} (e.g., the agent is at the red object). At every point in time, the CoS DNF matches the intention with the current sensory input. Upon a successful match, the CoS signals the completion of the EB and deactivates its intention.  The structure of EBs enables  segmentation of a continuous behavioral flow into discrete intentional (goal-directed) actions.

To represent the above, we've used coupled intention and CoS \textit{nodes}, linked to \textbf{perceptual} and \textbf{CoS} \textit{fields}.  An example of a perceptual field is one which takes camera input, and transforms it so that the y-axis represents maximum \textit{hue}, and the x-axis is pixel column~\cite{sandamirskaya2011neural}.  The corresponding CoS field, defined over the same axes as the perceptual field, serves as input to the CoS nodes.  Intention nodes provide top-down biases to the perceptual and CoS fields, and these biases effectively define the behaviors.  An intention node of a particular EB (e.g., ``find yellow'') will bias the appropriate hue in the perceptual field and the appropriate area (e.g., the center) of the CoS field.  Bottom-up input from the CoS field to an EB's CoS node allow the node to become active in response to the stimuli which define when the behavior has been completed ~\cite{sandamirskaya2011neural}.

Superposition of the perceptual field and the preshape from intention nodes results in regions of super-threshold activity, which then drive low-level motor commands via the motor field, e.g., setting an equilibrium point for a muscle or an angular velocity for the wheels of a mobile robot. An example motor field is a simple 1D space representing heading direction.  As the agent performs an action, environmental stimuli such as visual input from cameras, or position information from motor encoders, change continuously in time, resulting in changes in the pattern of activity across the perceptual field.

The intention nodes balance self-excitation, inhibition from its own CoS node, and excitation from its value node (value nodes are explained later).  The parameters are tuned so that, when no intention node is above threshold (sigmoidal $f$ is near zero for all) a winner-take-all behavior results.  Otherwise, a single intention node stays ``on'' (high $f$) due to self-excitation and suppression of the others.  The equation for each intention node's activity is given by:

\begin{IEEEeqnarray}{rCl}
\label{EQ:INTNODES}
\tau^{int} \dot{d}_i^{int} &=& -d_i^{int} + h^{int} + c_{+}^{int} f_{S}(d_i^{int}) + c_{val}^{int} d_i^{val}  \nonumber\\
&& -c_{-}^{int}\sum_{k \neq i} f_{S}(d_k^{int}) -c_{cos}^{int} f_{S}(d_i^{cos})
\label{eq:intention}
\end{IEEEeqnarray}

The CoS nodes signal when a behavior has been completed, on the basis of bottom-up perceptual input. The equation for each CoS node is given by:

\begin{IEEEeqnarray}{rCl}
\label{EQ:COSNODES}
\tau^{cos} \dot{d}_i^{cos} &=& -d_{i}^{cos} + h^{cos} + c_{+}^{cos} f_{S}(d_i^{cos})  \nonumber \\
&& -c_{int}^{cos} \sum_{k \ne i} f(d_k^{int}) + c_{input}^{cos} \sum_j f(U_{i,j}^{cos})
\label{eq:CoS}
\end{IEEEeqnarray}

Activities of both nodes,  ($d_i^{int,cos}$), in the absence of excitatory or inhibitory inputs are driven by a resting level, $h^{int,cos}$ as well as a passive decay term, $-d_i^{int,cos}$ which drives the node's activity back towards a resting equilibrium. Self-excitatory feedback ($c_{+}^{int,cos} f_{S}(d_i^{int,cos})$) stabilizes activity of a node if an external input pushes it through the activation threshold. Lateral inhibition ($-c_{-}^{int}\sum_{k \neq i} f_{S}(d_k^{int})$) among intention nodes causes these nodes to compete in a winner-take-all fashion, such that only a single intention node can remain on, while suppressing others. This competition is biased by nodes which encode learned values via the term $c_{val}^{int} d_i^{val}$. Unlike the intention nodes, the CoS nodes receive bottom-up inputs from the perceptual field $ c_{input}^{cos} \sum_j f(U_{i,j}^{cos})$ which excite a CoS node when environmental conditions match the expected context which defines that a behavior is completed.
Once a behavior is completed, the CoS node of the given behavior will become active, and shut down the active intention node by inhibitory inputs $-c_{cos}^{int} f_{S}(d_i^{cos})$.

In our simulation, we set the parameters in these equations as $\tau^{int}=\tau^{cos}=.3$, and $h^{int}=h^{cos}=5$. The inhibitory coefficients were set to $c_{-}^{int}=10$ and $c_{cos}^{int}=5$ and $-c_{int}^{cos}=2$, while the excitatory coefficients were set as $c_{+}^{int}=10$ and $c_{val}^{int} d_i^{val}=20$.

The sigmoid function $f_{S}$ ensures that output activations are bounded between 0 and 1, and is given by:

\begin{IEEEeqnarray}{rCl}
\label{EQ:FSIGMOID}
f_{S}=\frac{1 + \beta (x-\mu)}{2(1 + \beta |x-\mu|)}
\end{IEEEeqnarray}

Because of winner-take-all (WTA) competition between intention nodes of the EBs, only a single behavior can be selected and active at any given time. This competition is driven either by 1. endogenous random activity (during exploration), or 2. by long-term memory representations of  \textit{values} (during exploitation). These values, stored in weights $W_{ij}$, can be read out into value nodes.  In the absence of randomized exploration, the value weights specify a chain of behaviors.  They cause one behavior to reliably follow another.  Ideally, the chain of behaviors will serve to maximize the agents expected future reward.

The activity of the value nodes is computed as:

\begin{eqnarray}
\label{EQ:VALNODES}
\tau^{val} \dot{d^{val}_i}(t) &=& \sum_j f(d_j^{cos}) W_{ij}.
\end{eqnarray}

Afterwards they are divisively normalized to sum to one.

The value nodes, intention nodes, CoS nodes, perceptual and motor fields work together to produce a sequence of elementary behaviors.  In the next subsection, we discuss the RL part, the goal of which is to tune the values.

\subsection{Reinforcement Learner}

The second major component of DN-SARSA($\lambda$) is the reinforcement learner. An initial requirement of an RL system is a representation of states and actions.

\textbf{State-Action Representations.} In DN-SARSA($\lambda$) a state/action field is a set of discrete nodes organized in a matrix,  wherein each row receives input from one of the intention nodes, and each column receives inputs from one of the CoS nodes of the available EBs. The sites in the state/action field are excited in response to coincident activations of CoS and intention nodes, which happens only in a transition phase between two EBs. By detecting transitions in this manner, the \emph{states} in the RL sense are defined by the CoS nodes (i.e., which behavior the agent has just finished), and the \emph{actions} are defined by the intention nodes (i.e., what behavior the agent selects next).

The SA (state-action) cells ($I_{ij}$) are not implemented as differential equations, but rather assume steady state dynamics, and are defined by:

\begin{IEEEeqnarray}{rCl}
I_{ij}=[\sum_{k\ne i}\sum_{j\ne l}d_k^{int}d_l^{cos}] f_H(f_s(d_i^{int})f_s(d_j^{cos}))
\label{eq:TPCells}
\end{IEEEeqnarray}

\noindent where $f_H$ is the Heaviside step function.


\textbf{Transient Pulse (TP)-Cells.}  The activity within the state/action field excite another field of nodes known as transient pulse cells \cite{rhodesThesis}. Each node in this field is modeled as a coupled circuit composed of an excitatory and inhibitory TP cell ($TP^+$ and $TP^-$ respectively). The activities of each of the $TP^+$ cells in these circuits behave as onset and offset detectors for their respective state/action nodes, by producing a transient excitatory pulse in response to the onset of input from the state/action field, and a transient inhibitory (negative) pulse in response to the offset of that activity.


The behavior of the field of coupled excitatory ($TP_{ij}^+$) and inhibitory ($TP_{ij}^-$) cells is given by:

\begin{IEEEeqnarray}{rCl}
\tau^{TP}\dot{TP_{ij}^+}&=&(-TP_{ij}^+ + I_{ij} - TP_{ij}^-) \\
\tau^{TP}\dot{TP_{ij}^-}&=&(-TP_{ij}^- + I_{ij})
\label{eq:TPCells}
\end{IEEEeqnarray}

Both the excitatory and inhibitory cells contain a passive decay term ($-TP_i^+$ and $-TP_i^-$), as well as excitatory input from their corresponding state/action cells, $I_{ij}$. In addition, the $TP_{ij}^+$ receive inhibition from their corresponding inhibitory cell, $TP_{ij}^-$. For each intention, $i$, and each CoS, $j$, both cells ($TP_{ij}^+$ and $TP_{ij}^-$) are initially at rest. When the  input, $I_{ij}$, from the state/action field turns on, both cells integrate activity at a rate proportional to this input. However, whereas the $TP_{ij}^-$ cell integrates activity until it reaches equilibrium (while input remains on, equilibrium is reached at the value of the input), the $TP_{ij}^+$ cell will begin to decrease in activity as $TP_{ij}^-$ increases. In fact, it is easy to see that at equilibrium, $TP_{ij}^+=I_{ij} - TP_{ij}^-$, which will therefore approach zero. Once input shuts off, $TP_{ij}^+$ is approximately 0, whereas $TP_{ij}^-$ is approximately equal to the input strength. As a result, $TP_{ij}^+$ will experience an initial burst of inhibition, until both $TP_{ij}^+$ and $TP_{ij}^-$ then relax back to rest at 0. In both equations, the parameter $\tau=1/2$.

In DN-SARSA, the onset and offset detection capabilities of TP cells have multiple uses. Firstly, because they exhibit a fixed-width (in time) pulse of activation, they allow buffering of inputs to the eligibility trace layer, in order to prevent persistent inputs to those cells. Secondly, as consequence of the fact that they detect onsets and offsets of inputs, they can serve as the mechanism by which calculation $Q(s',a')-Q(s,a)$ is calculated. That is, if inputs occur in back-to-back fashion, such a mechanism results in the positive activation of TP cells corresponding to the currently active state/action pair $(s',a')$, while simultaneously producing negative activation of the previous state action pair $(s,a)$.

Activity from the $TP^+$ cells serves as input to a neural structure, wherein eligibility traces for the history of the activated state/action pairs is maintained, as described next.

\textbf{Eligibility Trace.}  Since the eligibility trace in RL \cite{sutton1998reinforcement} may be interpreted as a  form of a working memory, we simulate the eligibility trace (ET) field as an Item and Order working memory, which has been used to model a range of behavioral and psychological data regarding working memory, speech perception, and unsupervised sequence learning \cite{grossberg2011neural, grossberg1978WM}. Item and Order working memories encode the order of a sequence of presented items by the relative levels of activation across those items. In DN-SARSA($\lambda$), more recently occurring state/action transitions result in higher levels of activity in the ET field than those state/action transitions having occurred further in the past. This property emerges naturally due to a ubiquitous neural architecture, known as a recurrent on-center, off-surround network, whose cells obey shunting dynamics. This structure ensures that the summed total activity is bounded, and that shunting dynamics lead to divisive normalization,  which causes individual cell activities to be reduced by constant ratio factors upon presentation of new items. For a more technical analysis, see ~\cite{grossbergMemTheory}. Because of the recurrent on-center, off-surround structure, cell activities can reach sustained equilibrium values in the absence of inputs. Further, because the inputs to this field are brief duration pulses corresponding to the onsets of inputs from state/action representations, the activity pattern across this field reaches equilibrium, and is no longer altered regardless of how long the state/action cell itself remains active. Taken together, these processing capabilities give rise to a system which can sustain a fixed activation level as variable length actions are undertaken, and whose activities self-stabilize in periods between, as well as during, subsequent actions.

For a working memory cell which encodes the state/action pair indexed by $(i,j)$, its activity $u_{ij}$ is given by:

\begin{IEEEeqnarray}{rCl}
\tau^u \dot{u_{ij}}&=& (1-u_{ij})(\alpha f_P(TP_{ij}^+) + \beta u_{ij}) \nonumber\\
&& -u_{ij}(\alpha \sum_{k,l\neq i,j}f_P(TP_{kl}^+) + \beta \sum_{k,l\neq i,j} u_{kl})
\label{eq:ETField}
\end{IEEEeqnarray}

The cell's activity is bounded below by $0$, and bounded above by $1$ due to the excitatory shunting term $(1-u_{ij})$, which prevents the inputs from having any effect once $u_{ij}=1$, and the inhibitory shunting term $(-u_{ij})$ which prevents the inhibitory inputs from having any effect once $-u_{ij}=0$. Inputs from state/action pairs $(I_{ij})$ are pulse inputs resulting from joint activations in CoS and Intention nodes across the EBs. There are also on-center ($ \beta u_{ij}$) and off-surround ($\beta \sum_{k,l\neq i,j} u_{kl}$), which, when coupled with shunting dynamics, give rise to the Item and Order properties discussed above. The parameters are set as $\alpha=1.1$ and $\beta=.8$.


\begin{figure*}[!tp]
\centering
\includegraphics[width=0.86\linewidth]{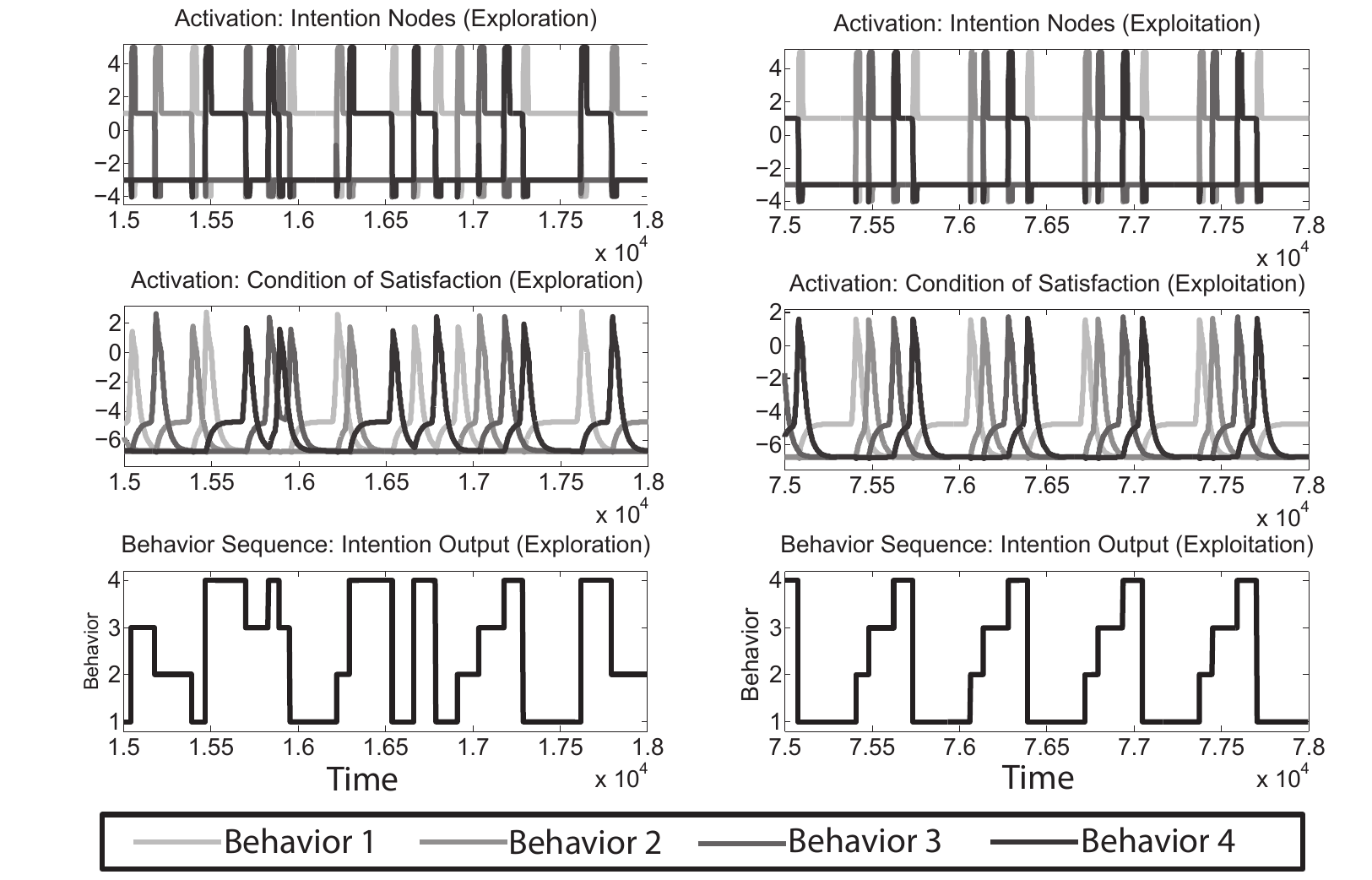}
\caption{Illustration of how our continuous time process model DN-SARSA($\lambda$) converts sensory-motor representations to discrete-like events.  Left: the activation of intention and condition of satisfaction nodes during a short chunk of time during the robot's exploration phase.  Left bottom: the intention node output, indicating the behavioral sequence.  Not all behaviors take the same amount of time.  Right: after learning.  The optimal sequence was learned and has stabilized. }
\label{FG:DYNAMICS}
\end{figure*}

\textbf{Value Opposition Field.}  The pattern of activity which unfolds across the eligibility trace field excites a value opposition (VO) field, which  prepares the calculation of the TD-error. In the VO field, the representations of the currently active state/action pair (with value $Q(s',a')$) and the negative of the previously active state/action pair (with value $Q(s,a)$) become active. This results from the onset/offset detections of state/action pairs by the TP cells in the following way. When the state/action pair $(s', a')$ is selected to be performed, it's corresponding $TP^+$ cell emits a pulse of activity. At the same time, the previous, just finished state/action pair, $(s,a)$, has a $TP^+$ cell emitting a negative pulse of activity, since its corresponding state/action representation is the most recent one to have turned off. All other $TP^+$ cell activities remain zero.  Consequently, the onset / offset detectors simultaneously exhibit excitatory activation in the currently active state/action pair, with inhibitory activation in the previously active state/action pair. These TP-cell activations gate inputs from the eligibility trace field to the VO field. These inputs are also weighted by LTM traces  which represent the Q-Values. Together, these multiplicative inputs ensure that the  activity in the VO field represent $Q(s',a')$, and $-Q(s,a)$.

Activity in the Value Opposition field follows the dynamics:
\begin{IEEEeqnarray}{rCl}
\tau^{O}\dot{O_{ij}}&=& (-O_{ij} + \gamma f_H(u_{ij}) W_{vu}f_H(TP_{ij}^+) \nonumber\\
&&-f_H(u_{ij})W_{vu}f_H(-TP_{ij}^+),
\label{eq:ValueOppField}
\end{IEEEeqnarray}

where the function $f_H(w)$ is the Heaviside function. Because the only excitatory $TP^+$ cell activity corresponds to the presently active state action pair, $(s',a')$, and the only inhibitory $TP^+$ activity corresponds to the previously active state action pair, $(s,a)$ at equilibrium gives $O_{ij} = (\gamma W_{(s',a')}-W_{(s,a)})$.


where the weights correspond to our learned Q-values, and the indices i,j have been replaced by the presently and previously active state action pairs. Our parameter $\tau^O=1/10$, and $\gamma=.8$.

\textbf{TD-Error.}  The TD-error is calculated in part by a value cell that receives excitatory inputs from all cells in the VO field. This ultimately results in a cell whose activity computes the difference between the stored LTM values for the currently and previously active State/Action pairs.

The value cell activity is given by:

\begin{IEEEeqnarray}{rCl}
\tau^v\dot{v}&=&(-v + \sum O_{ij})
\label{eq:TDError}
\end{IEEEeqnarray}

Because the LTM weights $W_{vu}$ ultimately come to encode our desired Q-values, the value cell at equilibrium calculates, $v = \sum_{ij} O \approx \gamma W_{Q(s',a')} - W_{Q(s,a)}$. The value stored here then modulates our learning law along with incoming rewards.

\textbf{LTM Weights (Q-Values).}  The update rule for the weight values of connections between the state/action pairs essential mirror the form of the update equation in SARSA. In particular, the  Q-values  (that is $Q(s',a')$ and $Q(s,a)$ representing the Q-values of the current and previous s/a pairs) are  values of the weights, and the  working memory based eligibility trace values correspond to the SARSA  eligibility trace values. The weight update equation is Eq.~\ref{eq:LTMWeights},

\begin{IEEEeqnarray}{rCl}
\dot{W_{ij}}&=&\alpha (1-SA_{ij})[r + v]u_{ij}
\label{eq:LTMWeights}
\end{IEEEeqnarray}

The summed activity across the VO field, plus any external reward present,  modulate the weights storing Q-values, as do the eligibility traces which are the pre-synaptic cells to these weights.


\begin{figure*}[!tp]
\centering
\includegraphics[width=\linewidth]{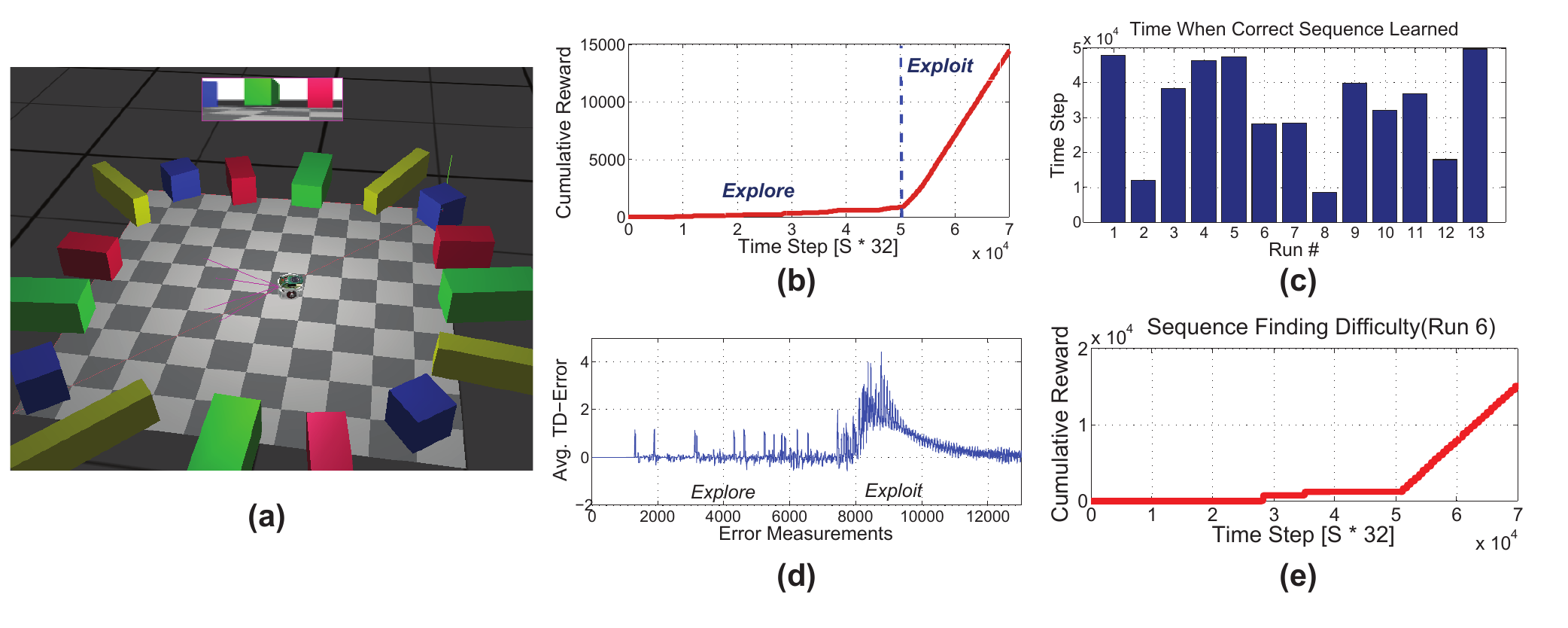}
\caption{
 (a)  Simulation environment in which a e-puck vehicle at the center rotates on the spot to direct its camera at colored objects and is rewarded for doing so in a particular order of colors.
 (b)  Cumulative reward as a function of time averaged across 13 runs.  In the first $50,000$ time steps (32 time steps per second), the system randomly selects intended colors; thereafter it selects the most valuable intended color.
 (c) Time needed to learn the rewarded sequence in each of the 13 runs.  (d)  Average TD-error.  (e)  The cumulative reward from example run (6)
 \vspace{-.1cm}}
\label{FG:RESULTS}
\end{figure*}

\section{Implementation and Results}
\label{SE:RESULTS}

\subsection{Environment and Behaviors}
The model is tested on a robotic vehicle simulated in the Webots simulator, performing a search for rewarding sequences of colored blocks, as illustrated  in Fig.~\ref{FG:RESULTS}(a).  The E-Puck robot is surrounded by 16 blocks of four different colors (red(R), green(G), blue(B), yellow(Y)), which are picked up by the robot's camera and are represented as localized color-space distributions in the perceptual DNF.  The robot ``finds'' a particular color, as determined by the currently active intention node, by rotating on the spot so that an object of the given color falls onto the center of the image of the vehicle's camera. Once centered, activation in the CoS node of the particular EB initiates a new EB to be performed (i.e., a new color to be searched for). If the robot finds the correct five-item sequence $G \rightarrow B \rightarrow Y \rightarrow R \rightarrow G$, a positive reward is provided for a few time steps.

We note here that the proof-of-concept implementation described in this section uses rather simplified EBs.  The DN-SARSA($\lambda$) learning system will work with other EBs than these ones, for example, more sophisticated DNF behaviors that enable mobile robots to deal with obstacles~\cite{bicho1997dynamic}.

Note that this is a POMDP, since our agent's state encodes the previously completed behavior only.  In our environment, the optimal policy is not representable given just the observable state. If we use TD(0), for example, the horizon will be too short --- if $R \rightarrow G \rightarrow Y \rightarrow B \rightarrow R$ is uncovered and rewarded, the model will first boost values from $B \rightarrow R$, and will next boost values of any of the three $R \rightarrow B$, $G \rightarrow B$, $Y \rightarrow B$, but there will be no feedback so that \textit{only} the correct one could be learned.  Memory of the last three behaviors is needed to reliably predict reward.  Due to the eligibility trace, DN-SARSA($\lambda$) can learn the sequence succesfully.  It is known that eligibility traces are not a complete solution to POMDPs, but eligibility traces can lead to good or even optimal POMDP solutions in some cases.

\subsection{Setup of the Model}

Initially, the value-encoding weights of the reinforcement learner are set to zero. Ultimately, the goal for the robot is that it discovers and learns the target sequence by reinforcement learning. We use a random exploration strategy during the first $50,000$ time steps in which noise is added to the weights. This causes the robot randomly select EBs for approximately $300$ orientation behaviors that occur during this period.  One could imagine future work using more sophisticated exploration methods~\cite{schmidhuber2010formal}.  After $50,000$ steps, the noise is removed and the robot operates in exploitation mode, consistently excecuting what it estimates to be its most valuable next behavior while continuing to learn.

\subsection{Results}

Please see Fig.~\ref{FG:DYNAMICS}.  This illustrates how temporally discrete events emerge from continuous time activation dynamics in the elementary behaviors. These events arise from instabilities in the neural dynamics triggered by CoS onsets. The left column illustrates the irregular activation of EBs during exploration, while the right column shows the consistent sequence of activated EBs in the exploitation phase.

Fig.~\ref{FG:RESULTS}(b)-(e) shows results in terms of the robot's learning performance.  In all trials in which the robot uncovered the rewarding sequence in exploration mode, it was able to eventually execute the optimal policy in exploitation mode. In some trials, the optimal policy was attained only in the exploitation phase, which showed that it is useful to maintain learning both during exploration and exploitation.  Learning in the exploitation phase consists primarily of \textit{unlearning} incorrect ``shortcuts'' inherited from the exploration phase.  This occurs, for example, when the robot finds the sequence, and correctly values the transition from $R \rightarrow G$ the most, but incorrectly also values the transition from any other color than $Y$ to $R$. During exploitation the robot realizes that shortcuts do not lead to reward (by executing them and not receiving any reward). Their values are diminished until the true rewarding sequence remains.

Fig.~\ref{FG:RESULTS}(c) shows the time at which the sequence was first uncovered. Fig~\ref{FG:RESULTS}(e) illustrates the reward from one run, in which the robot finds the target sequence a first time after about $30,000$ steps, finds it again (by luck). When the system enters exploitation mode its starts maximizing reward by doing the correct thing over and over again until the simulation ends.  Fig.~\ref{FG:RESULTS}(d) shows the averaged TD-error, illustrating that the neural system learns to predict discounted future reward.  The detection of reward acts as an instability for the reinforcement learner, and the learning mechanism is simply a constant drive towards stability. 

\subsection{Transfer to Real Robot}

\begin{figure}[!t]
\centering
\includegraphics[width=0.75\linewidth]{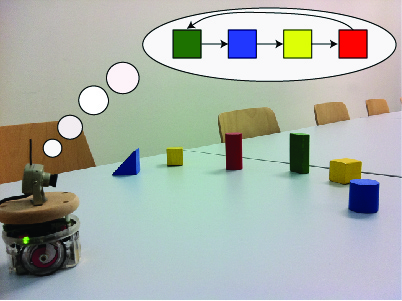}
\caption{The E-puck in its environment, surrounded by the colored objects of different sizes and shapes. The "thought bubble" shows the rewarding sequence of colors.}
\label{FG:ROBOT}
\end{figure}

To show that our system can deal with real sensory information and real motor system,  we transferred a set of weights learned from a successful run of simulation to a real E-puck (see Fig.~\ref{FG:ROBOT}). A video of the robot successfully moving through two iterations of the sequence is available at \url{http://www.idsia.ch/~luciw/videos/DFTBot.mp4}.

In the video, the top row shows the sensorimotor process: from sensory input to the perceptual field and to the motor field.  One can see the different colors that are detected along the hue dimension (Y-axis of perception), and how priming from the different intention nodes causes selection of one color and execution of the corresponding behavior.  Observe that the system is robust against perceptual noise and fluctuation in the visual channel (e.g. changing lighting conditions, shades, mismatch between the robotic and the simulated camera).  The activities of the intention and CoS nodes in the bottom row show the behavioral switching dynamics.  The CoS field is also shown here, which illustrates the link from perception to behavior completion.  Finally, the learned value weight matrix is shown, where white indicates a high value, with CoS (state) on the y-axis and intention (action) on the x-axis. Note that it encodes the rewarding sequence.

The successful transfer onto a real robotic system shows that the DN-SARSA($\lambda$) reinforcement learner brings about a representation that is capable of producing behavior in the physical robot based on continuous (raw) visual input and physical motors, driven by continuous-time dynamics.




\section{Conclusion}
\label{SE:CONCLUDE}

The DN-SARSA($\lambda$) model provides a framework which shows how computational learning algorithms can be incorporated into a continuous neural-dynamical model.  This enables autonomous learning and acting in continuous and dynamic environments, a challenge that is easily overlooked when formalizing the learning problem in discretized spaces without accounting for their coupling to sensory-motor dynamics.  Future work involves improving the exploration phase, which in this paper is a simple random action selection, and integrating with a recently introduced architecture for organizing elementary behaviors~\cite{richter2012robotic}.  DN-SARSA($\lambda$) can potentially lead to \textit{learning} of the behavioral constraints in that architecture.

\textbf{Acknowledgement.} We'd like to thank Alexander F\"orster for mounting the camera on the E-puck.

\label{SE:bib}


\end{document}